# RASEC: Rescaling Acquisition Strategy with Energy Constraints under SE-OU Fusion Kernel for Active Trachea Palpation and Incision Recommendation in Laryngeal Region

Wenchao Yue, *Graduate Student Member, IEEE*, Fan Bai, Jianbang Liu, *Graduate Student Member, IEEE*, Feng Ju, *Member, IEEE,* Max Q-H Meng*, Fellow, IEEE,* Chwee Ming Lim and Hongliang Ren, *Senior Member, IEEE*

*Abstract-* **A novel palpation-based incision detection strategy in the laryngeal region, potentially for robotic tracheotomy, is proposed in this letter. A tactile sensor is introduced to measure tissue hardness in the specific laryngeal region by gentle contact. The kernel fusion method is proposed to combine the Squared Exponential (SE) kernel with Ornstein-Uhlenbeck (OU) kernel to figure out the drawbacks that the existing kernel functions are not sufficiently optimal in this scenario. Moreover, we further regularize exploration factor and greed factor, and the tactile sensor's moving distance and the robotic base link's rotation angle during the incision localization process are considered as new factors in the acquisition strategy. We conducted simulation and physical experiments to compare the newly proposed algorithm - Rescaling Acquisition Strategy with Energy Constraints (RASEC) in trachea detection with current palpation-based acquisition strategies. The result indicates that the proposed acquisition strategy with fusion kernel can successfully localize the incision with the highest algorithm performance (Average Precision 0.932, Average Recall 0.973, Average $F_1$ score 0.952). During the robotic palpation process, the cumulative moving distance is reduced by 50%, and the cumulative rotation angle is reduced by 71.4% with no sacrifice in the comprehensive performance capabilities. Therefore, it proves that RASEC can efficiently suggest the incision zone in the laryngeal region and greatly reduced the energy loss.**

*The experiment video demo is available at https://sites.google.com/view/rasec/ ( *Corresponding author: Hongliang Ren*)

H. Ren is with the Department of Electronic Engineering and Shun Hing Institute of Advanced Engineering, The Chinese University of Hong Kong (CUHK), Hong Kong 999077, and also with the Department of Biomedical Engineering, National University of Singapore, Singapore 119077 (e-mail: hlren@ieee.org ).

W. Yue is with the Department of Electronic Engineering and Shun Hing Institute of Advanced Engineering, The Chinese University of Hong Kong (CUHK), Hong Kong 999077, and also with the Department of Biomedical Engineering, & Department of Mechanical Engineering, National University of Singapore, Singapore 119077 (e-mail: wenchao.yue@link.cuhk.edu.hk).

B. Fan and J. Liu are with the Department of Electronic Engineering, The Chinese University of Hong Kong (CUHK), Hong Kong 999077, (e-mail: fanbai@link.cuhk.edu.hk; henryliu@link.cuhk.edu.hk).

F. Ju is with the College of Mechanical and Electrical Engineering, Nanjing University of Aeronautics and Astronautics, Nanjing 210016, China (e-mail: juf@nuaa.edu.cn).

Max Q.-H. Meng is with the Department of Electronic and Electrical Engineering, Southern University of Science and Technology, Shenzhen 518055, China, with the Department of Electronic Engineering, The Chinese University of Hong Kong, Hong Kong, and also with the Shenzhen Research Institute of the Chinese University of Hong Kong , Hong Kong. (e-mail: max.meng@ieee.org )

C. M. Lim is with the Department of Otolaryngology-Head and Neck Surgery, Singapore General hospital and Duke-NUS Graduate Medical School, Singapore. (e-mail: chweeming.lim@gmail.com )

## I. INTRODUCTION

With the current increasing infections of COVID-19, respiratory failure becomes a frequent cause of mortality, in which 25.9% of pneumonia required admitted to Intensive Care Unit (ICU), and 20.1% developed acute respiratory distress syndrome (ARDS) [1-4]. Tracheotomy creates an artificial airway with a transverse incision between the trachea cartilages and plays a major role in buying time for mechanical ventilation [5]. With the development of robotic minimally invasive surgical systems (RMIS), robotic tracheotomy surgery has been further promoted [6, 7]. Prior to robotic tracheotomy surgery, precisely identifying the trachea rings and thus the appropriate localization of the incision region is a critical issue. In the traditional tracheotomy surgery, the surgeon finds the suitable incision location through palpation, usually from the inferior border of the thyroid cartilage to near the superior sternal fossa area. The localization accuracy highly depends on the surgeon's experience. However, the robotic tracheotomy system hardly determines the appropriate position of the incision or trachea rings by direct palpation without tactile feedback directly [8]. Tactile sensors can be installed at the tracheotomy robot's distal end for tactile feedback to the surgeons [9]. We aim at an RMIS system for tracheotomy that recognizes trachea rings and enables incision localization based on existing tactile sensors as in [10-12].

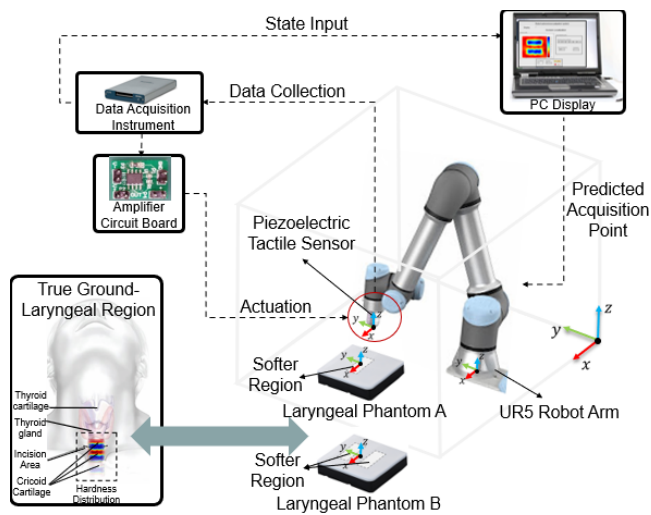

Figure 1. The overall experimental setup and potential application for active trachea palpation, and the visualization of the hardness distribution can help the surgeon recommend the most suitable location for laryngeal intubation.

There are under-explored aspects in the current tactile localization research: the lack of research on incision localization in a robotic tracheotomy, the homogeneous form of the kernel function in tactile localization situations, and imbalance impacts of each component through iterations in the acquisition strategies. Therefore, the paper aims to put forward a novel palpation-based strategy with energy constraints for trachea exploration and localization and then suggest the most suitable incision area for robotic tracheotomy. The main contributions can be summarized in our work as:

- Introduce tactile exploration and localization framework in the scenario of robotic tracheotomy surgery, targeting the problem of trachea localization and incision detection, filling a gap in this area of research.

- Propose a novel fusion kernel through combination Squared Exponential (SE) kernel with Ornstein-Uhlenbeck (OU) kernel to optimize the fitting smoothness of the model compared to the widely used kernel function.

- Design a new rescaling acquisition strategy with energy constraints, which normalizes each strategy term and considers the energy constraints, greatly improving acquisition efficiency and energy utilization.

## II. Related Work

The current research on active trachea palpation and incision recommendation are quite rare. Similar research can refer the lesion localization through the tactile method. There are mainly two methods to localize the target region using tactile sensors: construct tactile sensing arrays to detect the target tissue or uses tactile sensors to realize discrete measurement for perceiving the overall hardness distribution [13]. However, as for the tactile sensing arrays, the accuracy of the estimation hardness distribution is related to the array density, which puts higher requirements on the manufacturing process. Also, the tactile sensing array presses hard on the tissue surface, which would generate obvious deformation and might cause pain to the patient. While as for the discrete measurement, the hardness distribution can be obtained after measuring the hardness value at each point. Although the point-by-point measurement method can obtain the complete hardness distribution map of the target area, the time cost and energy cost during the tactile localization will be very high. Therefore, to reduce the high distance, rotation angle and time cost of the acquisition point by point, an efficient acquisition strategy is introduced to actively select the point to measure in the palpation process to achieve efficient localization and shape detection of the target region [14].

To reduce the localization time, Gaussian Process Regression (GPR) approach is applied to estimate the hardness distribution of the target region. Combined with Squared Exponential (SE) Kernel, continuous sweep sampling was proposed to estimate the tissue surface's geometry and hardness based on the GP model [15]. Therefore, the measured target geometry could be aligned with the preoperative model. Later an online update method was proposed to improve geometry detection error from 1.3676 mm to 1.3203 mm [16]; differently, to realize the hardness distribution estimation, the discrete sampling method with SE kernel was used under the GP framework. The acquisition strategy chose Expectation Improvement (EI) to balance exploration term and greed term, achieving the RMSE at the range of 0.74114 mm [17]. The Expectation-Maximization (EM) strategy was utilized under the Markov random field, sampling preferably at the target center with the highest evaluation [18]. However, this approach hardly explores the lesion boundaries and ignores the potential presence of other target areas in the unknown area. To localize the lesion boundaries, Implicit Level Set Upper Confidence Bound (ILS-UCB) algorithm combined with SE kernel gives the acquisition priority to the point with the greatest uncertainty in the estimated lesion boundary directly [19]. However, this strategy still cannot solve the problem of multi-target detection. Active Area Search (AAS) and Level Sets Estimation (LSE) algorithms were applied to simultaneously determine the shapes of tissue abnormalities under the multiple target regions scenario [20]. Later, another combination pattern with discrete acquisition and continuous segmentation was proposed to improve the efficiency [21], as summarized in Table I. Gaussian Process Implicit Surfaces (GPIS) instead of SE kernel was imported to describe the target shape. Their acquisition strategy collected information from failed grasp attempts and then attempted grasps in areas with the highest uncertainty (UNC) along the implicit surface to estimate the target contour [22]. When the target region is a narrowly closed surface, the widely used SE kernel function can lead to fitting failure due to its excessive smoothness; to guarantee that the estimated incision shapes are closed regions, Thin Plate (TP) kernel instead of SE kernel was sometimes utilized to describe the correlation among points. However, TP Kernel sacrifices acquisition stability because of the poor smoothness [23, 24]. Therefore, the acquisition efficiency should be examined together with optimizing the acquisition strategy and the kernel function to ensure integrity.

TABLE I. Method Summary To Optimize Acquisition Strategies Regarding The Localization Of Target Region

| Authors & year | Sensing Device | Kernel Function | Acquisition Strategy |
|---|---|---|---|
| Chalasani *et al.*, 2016 | ATI Mini 40 F/ T Sensor | Squared Exponential (SE) kernel | Continuous Sweep Sampling |
| Ayvali *et al.*, 2016 | ATI Nano 43 6-axis Force Sensor | SE kernel | Expectation Improvement (EI) |
| Nichols *et al.*, 2015 | Force-Torque Sensor + Phantom Premium 1.5 | Markov random field | Expectation-Maximization (EM) |
| Dragiev *et al.*, 2013 | Noisy Tactile Sensor | Gaussian Process Implicit Surfaces (GPIS) | Uncertainty Sampling (UNC) |
| Garg *et al.*, 2016 | Disposable Haptic Palpation probe | SE kernel | Implicit Level Set Upper Confidence Bound (ILS-UCB) |
| Salman *et al.*, 2018 | ATI Nano 25 F/T Sensor | Corrected SE kernel with input uncertainty | Active Area Search (AAS) & Level Sets Estimation (LSE) |
| Chalasani *et al.*, 2018 | ATI Mini 40 F/ T Sensor | SE kernel | Online Estimation Method |
| Yan *et al.*, 2021 | Flexible Hall Tactile Sensor | SE kernel | EI + Continuous Segmentation |
| Our contribution, 2021 | Piezoelectric Tactile Sensor | SE-OU fusion kernel | Rescaling Acquisition Strategy with Energy Constraints (RASEC) |

## III. PROBLEM STATEMENT

In this paper, the incision location for robotic tracheotomy is often specified in the softer trachea ring between the certain cricoid cartilages, which will project a closed area on the tissue surface $\Gamma_{obs} \subset \mathbb{R}^2$ distinct from the surrounding tissue. Therefore, localizing the trachea ring can be equivalently transformed into the problem to estimate hardness distribution $S(x)$ on the surface of the selected laryngeal area. The hardness distribution of the laryngeal region is described through Gaussian processes (GPs). Generally, these GPs are defined by their mean values and covariance functions. The hardness distribution on the tissue surface is assumed to be a continuous function between the hardness value $y = [y_1, y_2, ..., y_n]$ and location $x = [x_1, x_2, ..., x_n]$, defined on a Riemannian manifold $f: x \to y$. Assume $x, x' \subset \Gamma_{obs}$ are located in the acquisition region. Therefore, the prior hardness distribution field $f \sim \mathcal{GP}(\mu, k)$ is mapped with mean function $\mu(x)$ and kernel function $k(x, x')$. Here, the prior mean function is set to be zero for simplicity. Furthermore, the next acquisition point can be selected: $x_t = \arg\max_{x \in \Gamma_{obs}} G(u_{t-1}(x), \sigma_{t-1}(x))$, where $G$ is the acquisition function. Assuming that the tracheotomy robot's end effector is equipped with the tactile sensor, when the tactile sensor contacts the observation areas $\Gamma_{obs}$ in laryngeal region, to measure the hardness value from the corresponding location. After obtaining a set of obtaining the training dataset, GPR algorithm can be imported to estimate new distribution at the next acquisition point: $p(y_t \mid y, x_t) \sim \mathcal{GP}(K_t(K + \sigma_n^2 I)^{-1} y, K_{tt} - K_t(K + \sigma_n^2 I)^{-1} K_t^T)$, where $K$ is the $n \times n$ covariance matrix whose elements $K_{ij}$ ($i, j \in [1, ..., n]$) are calculated by certain kernel functions, $\sigma_n^2$ represents Gaussian white noise in the observation process. Similarly, $K_t$ is a $1 \times n$ vector defined as $K_t = [k(x_t, x_1), ..., k(x_t, x_n)]$, and $K_{tt} = k(x_t, x_t)$. To sum up theoretically above, the proposed active palpation-based incision localization problem is formulated as follows,

$$x_t = \arg\max_{x \in \Gamma_{obs}} G[u_{t-1}(x), \sigma_{t-1}(x),$$
$$d(x_{t-1}, x_t), \alpha(\beta_{t-1}, \beta_t)] \quad (1a)$$
$$\text{s.t. } d(x_{t-1}, x_t) \leq M, \quad (1b)$$
$$\alpha(\beta_{t-1}, \beta_t) \leq N. \quad (1c)$$

where $x_t$ is the predicted acquisition point. The functions $d(\cdot)$ and $\alpha(\cdot)$ refer respectively tactile sensor's moving distance and the robotic base link's rotation angle. Hence, by enforcing constraints Eq. (1b) and (1c), the next optimal acquisition point should also respect the energy budget in the palpation process.

## IV. METHOD

In this section, to optimize acquisition efficiency and energy utilization for trachea region localization, we first discuss the various kernel effects, construct a fusion kernel, and propose a novel acquisition strategy from state-of-the-art acquisition methods. Compared to the existing works on active acquisition strategy, our acquisition strategy normalizes each impact term into the same scale and introduces energy constraints (moving distance and rotation angle). Following the baseline above, we will discuss and optimize the acquisition strategy of trachea palpation from two aspects: the optimization of kernel function and the improvement of the acquisition strategy. Both aspects are discussed in more detail in the subsequent sections.

### A. Fusion Kernel Function

Since the GPR algorithm uses a GP as a prior, the kernel function is highly relevant to the prior information of the data and involve the correlation description between two points $x$ and $x'$. Therefore, the prediction results closely depend on the kernel function. The popular kernel function introduced in the GPR algorithm is the Squared Exponential (SE) kernel (shown as Eq. 2) with smooth interpolation characteristics due to its infinite differentiability, and stable convergent performance. Normally, the SE kernel follows as:

$$k_{SE}(x, x') = \sigma_f^2 \exp\left(-\frac{\|x - x'\|^2}{2l^2}\right), \quad (2)$$

where $\sigma_f^2$ refers to the variance as the amplitude factor, and $l$ is the length scale factor. Both are hyperparameters used to adjust the width of the covariance matrix. However, SEC kernel with the smooth interpolation often results in overfitting. It thus fails to identify the incision boundary accurately.

To avoid the fitting failure, a TP kernel derived from 2D implicit surfaces is introduced as Eq. 3:

$$k_{TP}(x, x') = 2r^2 \log|r| - (1 + 2\log(R))r^2 + R^2, \quad (3)$$

where $r = \|x - x'\|$ refers to the Eulerian distance between two points, and $R = \arg\max(\|x - x'\|)$ denotes the furthest Eulerian distance in the training set. As TP kernel is based on the assumption that the detection target is a closed area in the acquisition space, it is well suited to estimating 2D shape. However, TP kernel describes the interpoint relationship at the expense of smoothness and fitting stability.

Therefore, the relationship between smoothness and fitting stability should be a tradeoff. We utilized the constructure property of the kernel functions [25], and we performed a dot product fusion of the SE kernel function and the OU kernel function (shown as Eq. 5) along with proportionality factors. Noted that Ornstein-Uhlenbeck (OU) kernel simply replace the quadratic Euclidean distance with an absolute distance, which makes the output no longer smooth in the Eq. 4:

$$k_{OU}(x, x') = \exp\left(-\frac{\|x - x'\|}{l}\right), \quad (4)$$

$$k_{SE-OU}(x, x') = \sigma_f^2 \exp\left(-\frac{\alpha \|x - x'\|^2}{2l_1^2} - \frac{\beta \|x - x'\|}{l_2}\right), \quad (5)$$

where $\alpha$ and $\beta$ are proportionality factors used for adjusting the kernel effect, and $l_1$ and $l_2$ are the length scale factors. In this work, we choose $\sigma_f = 1$, $l_1 = 4\ mm$ and $l_2 = 3.5\ mm$. This dot product combination fixes the problem of being too smooth with a single SE kernel while solving an unstable fit with a single TP kernel. To compare the performance of different kernel functions, the RASEC algorithm probes the square sample of $50mm \times 50mm$ respectively described by different kernel functions, and the size of the single trachea phantom is $30mm \times 10mm$. Here the number of acquisition iterations choose 60. Table II shows the $F_1$ scores, which indicates that the RASEC algorithm has the best performance under the SE-OU fusion kernel.

TABLE II. EVALUATION PERFORMANCE OF RASEC ALGORITHM UNDER DIFFERENT KERNEL FUNCTION

| Kernel Function | Iteration Number | $F_1$ score |
|---|---|---|
| SE  |    | 0.943 |
| TP  | 60 | 0.923 |
| OU  |    | 0.910 |
| SE-OU |  | 0.951 |

Further attention to the distribution contour of the covariance matrix in Fig. 2, SE kernel is highly concentrated at the diagonal and completely ignores the relationship among distant points, ensuring high fitting smoothness; in contrast, the covariance matrix distribution from the TP kernel differs from that of the SE kernel, and the covariance matrix distribution takes full account of the inter-point relationships across the acquisition space to ensure the fitting accuracy, but at the expense of smoothness; while, the width of the covariance matrix distribution from OU kernel will be narrower relative to that from TP kernel, but cover a wider area than that from SE kernel, at the expense of smoothness and fitting stability, but with a very comprehensive description of the points near the diagonal.

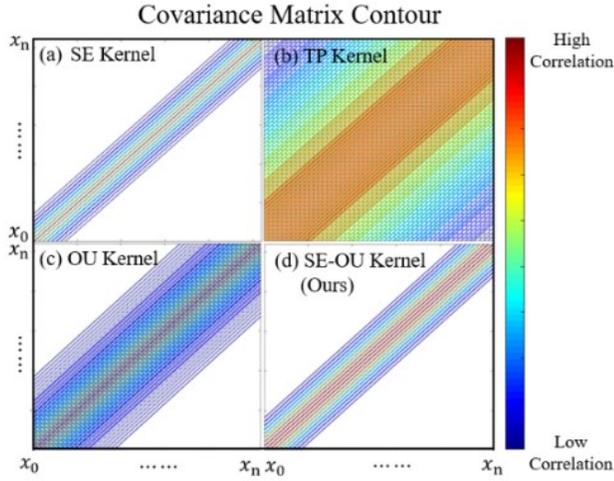

Figure 2. The comparison of covariance matrices contours generated by the (a) SE Kernel ($l$ = 4mm), (b) TP Kernel, (c) OU Kernel ($l$ = 4mm), and (d) our SE-OU Kernel ($l_1$= 4mm, $l_2$= 3.5mm), which demonstrates that the SE-OU kernel combines the advantages of the SE kernel's centralized description with the strengths of the OU kernel's integration consideration of points around the sub-diagonal of the covariance matrix.

The SE-OU kernel combines the advantages of the SE kernel's centralized description towards the acquisition region points with the strengths of the OU kernel's integration consideration around the sub-diagonal of the covariance matrix, ensuring both fitting stability and smoothness. Therefore, in this work, we choose SE-OU Fusion Kernel to describe the relationship amongst acquisition points.

*B. Rescaling Acquisition Strategy with Energy Constraints*

Active acquisition strategy for detection points is essential to realize the palpation-based incision localization. The design aim is to utilize the known hardness dataset to predict the next location of the optimal acquisition point through trading off greed factor $u(x)$ and exploration factor $\sigma(x)$. In this scenario, the greed factor $u(x)$ denotes estimated tissue softness, which means the greater $u(x)$ value is, the softer the estimated acquisition tissue is. Meanwhile, the exploration factor $\sigma(x)$ represents how much hardness information is known in a certain region. The greater $\sigma(x)$ value is, the less hardness information the acquisition tissue has. The existing popular acquisition strategies, such as Random Acquisition (Baseline), Uncertainty Sampling (UNC) [26], Expectation Improvement (EI) [21], Implicit Level Set Upper Confidence Bound (ILS-UCB) [27], Level Set Estimation (LSE) [28] are introduced to compare. Their hyperparameters is chosen based on the former study correspondingly. The efficiency of several acquisition strategies will be discussed and compared with our proposed RASEC. Considering the feature of these acquisition strategies: first of all, some strategy holds less possibility to localize multiple target regions like ILS-UCB, or not effectively localize multiple target regions, like UNC, LSE, EI. Moreover, their exploration term and greed term are not regularized into same level to weigh. The imbalance of both factors always occurs after acquisition iterations. Therefore, it is necessary to rescale these essential factors into the same scale. Additionally, regarding the energy minimization principle, the tactile sensor's total distance and robotic base link's rotation angle during palpation are not considered in current acquisition strategy when evaluating palpation efficiency. We focus on the base link because, among all robotic arm joints, the base link has the largest moment load in charge of the end-effector's weight, each upper linkage, the corresponding motor, etc. Therefore, reducing the base link's rotation angle can largely reduce the energy consumption in the trachea palpation process.

Therefore, to figure out the problems above, a novel acquisition strategy - RASEC is proposed shown in Eq. 6, considering the normalization of both essential parts, the moving distance of tactile sensor and the rotation angle of robotic base link, and the formula of RASEC is defined as:

$$x_t = \arg\max_{x \in \Gamma_{obs}} ((1-\theta)*\Phi_1(x) - \theta*\Phi_2(x) - K_d*\Phi_3(x) - K_\beta*\Phi_4(\beta)) \quad (6)$$

where,

$$\Phi_1(x) = \frac{\sigma_{t-1}(x) - \min_{x' \in \Gamma_{obs}} \sigma_{t-1}(x')}{\max_{x' \in \Gamma_{obs}} \sigma_{t-1}(x') - \min_{x' \in \Gamma_{obs}} \sigma_{t-1}(x')} \quad (7)$$

$$\Phi_2(x) = \frac{|\mu_{t-1}(x) - h_{t-1}| - \min_{x' \in \Gamma_{obs}} |\mu_{t-1}(x') - h_{t-1}|}{\max_{x' \in \Gamma_{obs}} |\mu_{t-1}(x') - h_{t-1}| - \min_{x' \in \Gamma_{obs}} |\mu_{t-1}(x') - h_{t-1}|} \quad (8)$$

$$\Phi_3(x) = \frac{\|x - x_{t-1}\| - \min_{x' \in \Gamma_{obs}} \|x' - x_{t-1}\|}{\max_{x' \in \Gamma_{obs}} \|x' - x_{t-1}\| - \min_{x' \in \Gamma_{obs}} \|x' - x_{t-1}\|} \quad (9)$$

$$\Phi_4(x) = \frac{\|\beta - \beta_{t-1}\| - \min_{\beta' \in B_{obs}} \|\beta' - \beta_{t-1}\|}{\max_{\beta' \in B_{obs}} \|\beta' - \beta_{t-1}\| - \min_{\beta' \in B_{obs}} \|\beta' - \beta_{t-1}\|} \quad (10)$$

Regarding the RASEC, there are four terms to mutually decide the localization of the next acquisition point $x_t$: the exploration term $\Phi_1(x)$, the greed term $\Phi_2(x)$, the distance term $\Phi_3(x)$ and the rotation angle term $\Phi_4(x)$. The $\Phi_1(x)$ term is used to evaluate the greed effect, and the $\Phi_2(x)$ term is used to evaluate the exploration effect in acquisition space. The third term $\Phi_3(x)$ and the fourth term $\Phi_4(x)$ are used to evaluate the tactile sensor's moving distance and robotic base link's rotation angle during palpation. Each term uses the Min-Max normalization to rescale into the same comparison level. Meanwhile, $\theta$ is the orthogonal factor, which holds the same meaning in ILS-UCB. $K_d$ and $K_\beta$ are respectively distance proportionality factor and rotation angle proportionality factor utilized for adjusting the tactile sensor's moving distance and robotic base link's rotation angle influence towards the acquisition strategy. The impacts of each term can be adjusted through the proportionality factor. Here, $\theta$, $K_d$ and $K_\beta$ are set as 0.5 to better balance each factor's weight. Its core idea is that after localizing the shape of a single trachea region, it will prefer to explore unknown regions to search for a potentially existing target. Since both the exploration and greed terms are normalized, both tradeoffs efficiency will be greatly improved. Meanwhile, considering the predicted acquisition points with similar boundary recognition effects (the weighted sum between the exploration term and greed term), multiple acquisition points differ in the tactile sensor's distance and base link's rotation angle referring to the current point. Considering the energy minimization, the third term and fourth term ensure that the robot arm selects the point more adjacent to the current state with the minimum moving distance cost and rotation angle cost.

## V. RESULTS AND DISCUSSIONS

In this section, we utilize the MATLAB platform for simulation. The effects of various acquisition strategies are compared and discussed according to the estimation hardness distribution results. An experimental study is further operated to verify the feasibility of the proposed RASEC algorithm. In the subsequent paragraphs, the simulation and experiment study are both discussed in detail.

### A. Simulation study

This simulation study discusses the effect of RAS without energy constraints on the integrated acquisition efficiency separately, and then focuses on the influence of moving distance term and rotation angle term. Additional scripts are imported to simulate the acquisition process towards the trachea region, which correspondingly simulates the trachea localization in the laryngeal region. The simulation sample is set as the square scale of $50mm \times 50mm$, and its origin is at the square center, within multiple tracheae similar to the human larynx structure. The size of the single incision region is $30mm \times 10mm$. The acquisition region is discretized and classified first: when the acquisition point $x$ locates inside the trachea region, its corresponding hardness value $y$ is classified to $\lambda$, and when $x$ locates outside the trachea region, $y$ is classified to $\gamma$. When $x$ locates on the trachea boundary, $y$ is classified as $(\lambda + \gamma)/2$. $\lambda$ value is always set higher than $\gamma$, because the region corresponding to $\lambda$ is the trachea region assigned higher expectations. The number of acquisition iterations is set at 60 in the virtual environment. To neutralize the error, we operate each acquisition strategy for the same sample mentioned above for ten acquisition trials, and take the arithmetic mean, and then compare among each other. The hardness estimation distribution maps of each acquisition strategy are shown in Fig. 3. It can be seen that, except for ILS-UCB, the rest of the acquisition strategies can successfully detect the presence of multiple trachea regions. Table Ⅲ shows the average precision rate, recall, and F1 scores of various acquisition strategies above in this simulation. It also indicates that the RASEC algorithm has better performance than other current acquisition strategies in trachea localization with an average precision of 0.932, an average recall rate of 0.973, and an average $F_1$ score of 0.952.

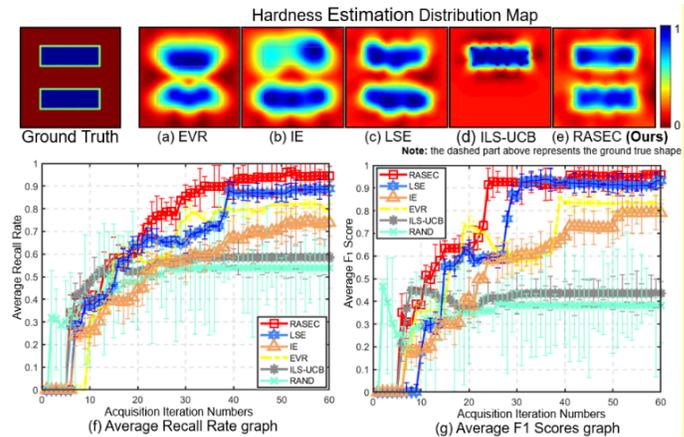

Figure 3. (a)-(e) show the estimation results of the sample hardness distribution after applying the five acquisition strategies, and the following figures (f) (g) demonstrate the average $F_1$ score and the average recall rate of the five acquisition strategies after ten sampling trials.

TABLE III. EVALUATION PERFORMANCE FOR DIFFERENT ACQUISITION STRATEGIES

| Acquisition Strategy | Average Precision | Average Recall | Average F₁ Score |
|---|---|---|---|
| Random Sampling-Baseline | 0.766 | 0.397 | 0.523 |
| EVR | 0.791 | 0.813 | 0.802 |
| IE | 0.780 | 0.786 | 0.783 |
| LSE | 0.867 | 0.918 | 0.892 |
| ILS-UCB | 0.911 | 0.432 | 0.586 |
| **RASEC-Ours** | **0.932** | **0.973** | **0.952** |

Palpation-based localization simulation is also performed for samples with multiple trachea settings to investigate the moving distance term and the rotation angle term in the RASEC. Since the robotic base link's rotation angle is directly related to the distance between the robot and the sample, here, the distance from the robot base to the center of the sample is assumed as $50mm$. Therefore, the proposed RASEC algorithm in this paper is not only better than other acquisition strategies for simultaneous localizing incision but also can highly optimize the moving distance of tactile sensor and rotation angle of the base link, thus shortening the palpation detection time and highly reducing the energy loss, meanwhile improving the detection efficiency.

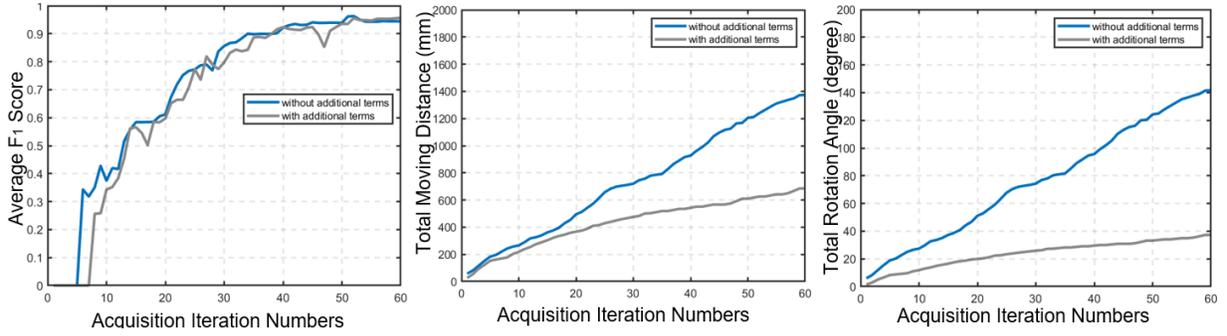

Figure 4. Comparison graph on the $F_1$ score, total moving distance and total rotation angle of proposed RASEC function with or without additional energy constrict term. This indicates that the cumulative moving distance is reduced by 50%, and the cumulative rotation angle is reduced by 71.4%.

The $F_1$ scores between RASEC algorithm with and without additional term tend to be same as the number of acquisition points increases (shown in Fig. 4), indicating that it can achieve accurate localization and shape detection of multiple trachea regions with or without adding the two energy constraint terms of moving distance and rotation angle. Regarding energy utilization, the tactile sensor's moving distance and the base link's rotation angle are greatly affected, especially the rotation angle of the base. The total moving distance is reduced to nearly 50%, and the total rotation angle is reduced to 71.4% after adding up the distance constraint term and rotation constraint term.

### B. Experiment study

The experiment platform is established to verify the capacity to locate the tracheal incision in the real sample, as shown in Fig. 5. To mimic the Robotic Tracheotomy, we use Ecoflex™ 00-10 to create actual samples corresponding to the laryngeal region according to the shape of the incision and the actual acquisition range (small range, large range) and use our tactile sensor installed on the UR5 robotic arm's end to measure hardness value with slight contact. This tactile sensor holds the threshold of 192 Hz/ 0.6 MPa, which can realize hardness measurement of soft tissue, and a hard range (which exceeds 0.6 MPa) with higher frequency to realize the identification of the stiff tissue. Therefore, this tactile sensor could meet our specific need for incision localization. Two silicone samples are with the size of $50mm \times 50mm$ detection area made with Ecoflex 00-10 as the phantom of the laryngeal sampling area. One had a groove in the base as a single incision, and the other had two grooves as a multi-incision distribution area. When the mixed Ecoflex™ 00-10 is injected into the base, the area corresponding to the groove will be softer, like the softer trachea area in the larynx region, as shown in Fig. 5 (b)(c). As for laryngeal phantom A& B, the number of acquisition iteration in the experiment is respectively set to 17 and 40. Considering the principle of energy minimization, the sensor's end in this letter contacts with the sample in a sliding-like manner to move in close proximity to the sample surface as shown in Fig. 5 (d)(e). The experimental results further verify that the proposed RASEC algorithm can achieve the objectives of incision localization and shape detection efficiently through active tactile acquisition.

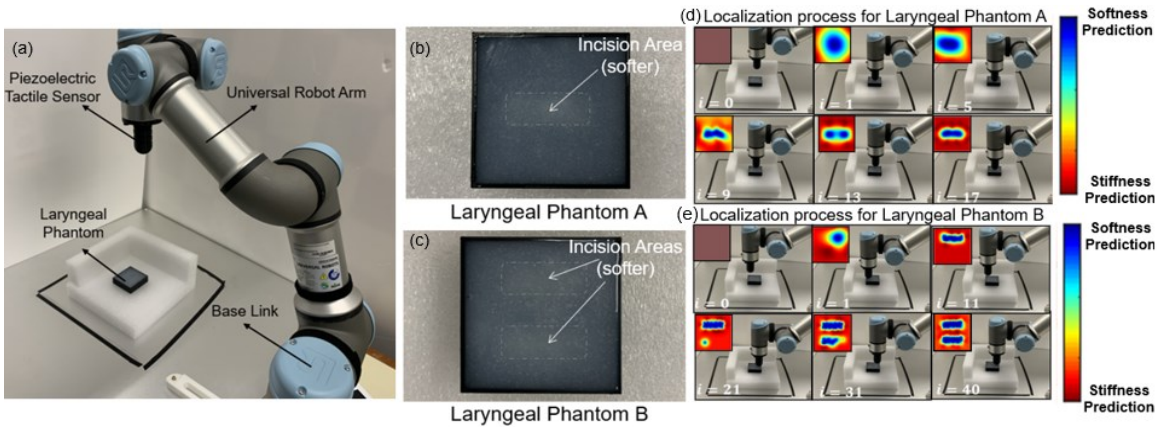

Figure 5. Experimental setup. (a) the experimental platform of the robotic autonomous palpation system for the laryngeal region; and (b) and (c) represents the Laryngeal Phantoms set with a single groove and multiple grooves. (d) and (e) describe the localization process for laryngeal phantom A & B: these two processes demonstrate that regarding the single incision case, the acquisition iteration number is ~17 times to localize the target area; while regarding the multiple incision case, the iteration number is ~40 times to realize localization.

## VI. CONCLUSION AND FUTURE WORK

This paper proposes an acquisition strategy for active trachea palpation and incision recommendation in the laryngeal region, with the improvement of kernel function and the design of the novel acquisition strategy. As for the kernel function part, based on the target of active trachea palpation and incision recommendation, the SE-OU fusion kernel is proposed according to the needs of this paper. The acquisition strategy- RASEC is further discussed and compared with the existing acquisition strategies. The simulation study proves that the proposed RASEC algorithm can achieve the localization of multiple incisions existing in the laryngeal region, shorten the tactile sensor's moving distance, and greatly reduce the base link's rotation angle, further improving the algorithm's efficiency. Finally, the experimental platform is established with a tactile sensor installed on the UR5 robot arm's end for palpation experiments. Two silicone samples with the groove-setup base are built as the laryngeal phantoms. The physical palpation experiment further proves the effectiveness of the proposed RASEC algorithm in trachea localization and incision recommendation.

The future work is to implant the visual alignment module to realize the coordinate calibration between the physical sample and the robot. At the same time, this paper lays a certain foundation for the subsequent robotic tracheotomy surgery. A more practical and well-established robotic system for tracheotomy surgery will be built by adding the control module onto the incision recommendation.

ACKNOWLEDGMENT

The authors would like to thank Yahui Yun and Yingxuan Zhang for their help understanding the GPR algorithm，Xinyi Zhou's suggestions on visualizing kernel functions, and Yinhe Li's contribution during the video processing.